\documentclass[runningheads,envcountsame,a4paper]{llncs}
\usepackage{graphicx}

\usepackage{amsmath}
\usepackage{booktabs}
\usepackage{algorithm}
\usepackage{algorithmic}

\usepackage{multirow}
\usepackage{array}
\usepackage{subfigure}
\usepackage{bm}
\usepackage{amsfonts}
\usepackage{color}

\usepackage[misc]{ifsym}

\begin{document}

\title{Inductive Unsupervised Domain Adaptation for Few-Shot Classification via Clustering}

\titlerunning{Inductive UDA for Few-Shot Classification via Clustering}

\toctitle{Inductive Unsupervised Domain Adaptation for Few-Shot Classification via Clustering}

\author{Xin Cong\inst{1,2} \and
Bowen Yu\inst{1,2} \and
Tingwen Liu\inst{1,2} (\Letter) \and
Shiyao Cui\inst{1,2} \and
Hengzhu Tang\inst{1,2} \and
Bin Wang\inst{3}
}

\authorrunning{X. Cong et al.}
\tocauthor{Xin Cong, Bowen Yu, Tingwen Liu, Shiyao Cui, Hengzhu Tang, Bin Wang}

\institute{Institute of Information Engineering, Chinese Academy of Sciences, Beijing, China \and School of Cyber Security, University of Chinese Academy of Sciences, Beijing, China \\
\email{\{congxin,yubowen,liutingwen,cuishiyao,tanghengzhu\}@iie.ac.cn} \\
\and Xiaomi AI Lab, Xiaomi Inc., Beijing, China \\
\email{\{wangbin11\}@xiaomi.com}}

\maketitle              

\begin{abstract}
Few-shot classification tends to struggle when it needs to adapt to diverse domains. 
Due to the non-overlapping label space between domains, the performance of conventional domain adaptation is limited.
Previous work tackles the problem in a transductive manner, by assuming access to the full set of test data, which is too restrictive for many real-world applications.
In this paper, we set out to tackle this issue by introducing a inductive framework, DaFeC, to improve \textbf{D}omain \textbf{a}daptation performance for \textbf{Fe}w-shot classification via \textbf{C}lustering.
We first build a representation extractor to derive features for unlabeled data from the target domain (no test data is necessary) and then group them with a cluster miner.
The generated pseudo-labeled data and the labeled source-domain data are used as supervision to update the parameters of the few-shot classifier.
In order to derive high-quality pseudo labels, we propose a Clustering Promotion Mechanism, to learn better features for the target domain via Similarity Entropy Minimization and Adversarial Distribution Alignment, which are combined with a Cosine Annealing Strategy.
Experiments are performed on the FewRel 2.0 dataset. Our approach outperforms previous work with absolute gains (in classification accuracy) of 4.95\%, 9.55\%, 3.99\% and 11.62\%, respectively, under four few-shot settings.

\end{abstract}
\section{Introduction}

\begin{table}[!htbp]
	\centering
	\begin{tabular}{l|l|m{0.28\textwidth}|l|m{0.28\textwidth}}
		\toprule
		& \multicolumn{2}{c|}{\textbf{Source domain: Wikipedia}} & \multicolumn{2}{c}{\textbf{Target domain: PubMed}} \\
		
		\midrule
		\multirow{5}{*}{Support}
		& \textit{member\_of} & \textcolor{red}{\textit{Newton}} served as the president of \textcolor{blue}{\textit{the Royal Society}}. & \textit{may\_treat} & \textcolor{red}{\textit{Ribavirin}} remains essential to \textcolor{blue}{\textit{Chronic Hepatitis C}} treatment. \\
		
		\cline{2-5}
		& \textit{instance\_of} & The \textcolor{red}{\textit{Romanian Social Party}} is a left \textcolor{blue}{\textit{political party}} in Romania. & \textit{classified\_as} &  All references about a \textcolor{red}{\textit{viral infection}} called \textcolor{blue}{\textit{ebola haemorrhagic fever}} were reviewed. \\
		\midrule
		
		Query & \textit{Which?} & \textcolor{red}{\textit{Euler}} was a member of \textcolor{blue}{\textit{the Royal Swedish Academy of Sciences}}. & \textit{Which?} & The \textcolor{red}{\textit{dental cysts}}, especially \textcolor{blue}{\textit{radicular cysts}}, are compared. \\
		\bottomrule
	\end{tabular}
	
	\vspace{1mm}
	\caption{ An example comes from FewRel 2.0, a few-shot dataset for relation classification with domain adaptation. Different colors indicate different entities, red for head entity, and blue for tail entity. Relation classification aims to determine the relation between two given entities based on their context.}
	\label{tab:example}
\end{table}

%
%
%
%
%
%

Few-shot classification aims to learn a classifier to recognize unseen classes with few labeled examples.
While significant progress has been made~\cite{vinyals2016matching,DBLP:conf/icml/FinnAL17,snell2017prototypical,DBLP:journals/corr/abs-1803-02999,DBLP:conf/cvpr/SungYZXTH18}, most previous works are under the assumption that the samples of unseen classes should be drawn from the same domain as the training data that was used to train the model.
However, in the real world, the application can be used in unusual environments and novel datasets, which means that these samples are likely from different domains.
Even a slight departure from a model’s training domain can cause it to make spurious predictions and significantly hurt its performance.
Table \ref{tab:example} illustrates a typical example of domain shift in the relation classification task which aims to classify the semantic relation between entities in a sentence.
The training data is collected from Wikipedia, but the actual data encountered at test time comes from PubMed, a biomedical literature corpus. 
The new relations in the target domain such as \textit{may\_treat} are different from those in the source domain.
Due to distinct domain characteristics like morphology and syntax, the performance of existing few-shot models drops drastically in such a situation. 

%
%
%
%
%
%

Unsupervised domain adaptation algorithms (UDA) aims at addressing the domain shift problem between a labeled source dataset and an unlabeled target dataset \cite{Ganin2015DomainAdversarialTO,DBLP:conf/cvpr/TzengHSD17,DBLP:conf/aaai/ZouZYLDS19}.
Conventional UDA methods typically assume that the target domain shares the same label space with the source domain so that the knowledge can be transferred from across domains via these same labels.
However, in the few-shot settings, the source domain and target domain do not have any overlap in categories. 
This unique setting renders most existing UDA methods inapplicable.
Previous work \cite{DBLP:conf/pkdd/DongX18} solves this problem by making use of test data (or query set in the few shot scenario) from the target domain in the transductive manner.
However, in some real-world scenarios, it is completely unrealistic to forecast test data in advance. 
In this paper, we work on a more realistic setting: inductive unsupervised domain adaptation for few-shot classification.
It is obvious that although we do not know the ground truth classes of the target domain, some of the unlabeled target-domain data may belong to the same classes. 
According to the cluster hypothesis \cite{chapelle2009semi}, the features of unlabeled data with the same latent label may cluster together in the representation space.
Mining these latent cluster structures can provide auxiliary information about the target domain, which could be beneficial to improve the adaption ability of few-shot models.
Based on such motivation, we design a novel framework named \textbf{DaFeC} (Unsupervised \textbf{\underline{D}}omain \textbf{\underline{a}}daption for \textbf{\underline{Fe}}w-shot classification via \textbf{\underline{C}}lustering), which effectively train the few-shot classifier with clustering-generated pseudo labels. 
The first step of DaFeC is the training of a representation extractor.
Based on the features of unlabeled target-domain data derived from the extractor, a cluster miner is applied to group these unlabeled instances and the subsequent cluster assignments are deemed as pseudo labels.
Finally, a few-shot classifier is trained based on both target-domain data with pseudo labels and source-domain training data to enable the classifier to adapt to the target domain.

%
%
%
%
%
%

Intuitively, the quality of pseudo labels significantly influences the performance of the few-shot classifier.
Theoretically, if input features are well discriminative, the cluster miner can group the instances easily and assign them pseudo labels with high-confidence.
Therefore, to generate high-confidence pseudo labels, we further propose a \textbf{C}lustering \textbf{P}romotion \textbf{M}echanism (\textbf{CPM}) to assist in training the representation extractor to produce cluster-distributed features for unlabeled target-domain data.
CPM contains three modules: 
%
%
First, to encourage features with the same latent class to get closer, we design a Similarity Entropy Minimization (CPM-S) objective. 
It calculates the Euclidean distance between each instance and others in the target domain and then minimizes the entropy of instance-wise distance vector to drive similar instances closer.
%
%
%
Second, in our preliminary study, we observe that although the source and target domain have different label space, they may still share some similar but not identical labels.
For instance (Table \ref{tab:example}), the class \textit{classified\_as} from the target domain is semantically similar to \textit{instance\_of} from the source domain with slightly difference.
Inspired by this phenomenon, we design an Adversarial Distribution Alignment method (CPM-A).
It introduces a domain discriminator to play an adversarial minimax game with the representation extractor to align the distribution of similar classes cross domains.
%
%
Third, we propose a Cosine Annealing Strategy (CPM-C) to support learning with CPM-S and CPM-A for achieving the optimal domain adaptation performance.

To summarize, our contributions are as following:

\begin{itemize}
	\item For the first time, we present an inductive unsupervised domain adaptation framework, DaFeC, for few-shot classification. To the best of our knowledge, there is no similar work in few-shot classification.  
	
	\item We propose a Clustering Promotion Mechanism to help the representation extractor produce cluster-distributed features for the generation of high-confidence pseudo labels.
		
	\item Our presented DaFeC is model-agnostic, which means that it can be incorporated into other models.
	
	\item Our approach achieves new state-of-the-art performance on FewRel 2.0, the currently largest unsupervised domain adaptation dataset for few-shot classification, delivering 3.99-11.62\% absolute gains over previous work\footnote{The source code and data of this paper are available now and they can be obtained from https://github.com/congxin95/DaFeC.}.
\end{itemize}

\section{Related Work}
Few-shot classification aims to develop models and algorithms which are able to recognize novel classes based on few labeled instances. 
Recently, meta-learning has been shown to be highly effective in few-shot learning, which can be generally classified into three categories: 
(1) Model-based methods \cite{santoro2016meta,DBLP:conf/icml/MunkhdalaiY17} design a special module such as memory to exploit meta information to make models generalize to new tasks rapidly with only a few instances.
(2) Optimization-based methods \cite{DBLP:conf/icml/FinnAL17,DBLP:journals/corr/abs-1803-02999} aim at learning a good initialized parameters which can achieve good performance through a few update steps. 
(3) Metric-based methods \cite{vinyals2016matching,snell2017prototypical,DBLP:conf/cvpr/SungYZXTH18} attempt to learn a good metric function which embeds data with the same classes into adjacent distance space.
Although many existing few-shot methods have achieved promising results, the performance of these methods is significantly degraded when the test data are drawn from different domains from training data, which is a quite common case in the real world.

Domain adaptation methods aim at exploiting labeled data in the source domain to perform a prediction task in the target domain.
Because annotating sufficient labeled data is time-consuming and labor-intensive, unsupervised domain adaptation (no need for labeled data of the target domain), has been extensively studied recently~\cite{Ganin2015DomainAdversarialTO,Tzeng2017AdversarialDD,DBLP:conf/aaai/ZouZYLDS19}. 
However, all of these methods assume that the categories of the target domain are shared with the source domain, which is too restrictive to generalize to the novel classes in the few-shot classification scenario.
To address this issue, \cite{DBLP:conf/pkdd/DongX18} leverages reinforcement learning to select source data similar to the target test data to train few-shot classifiers. 
Nevertheless, this method works in the transductive manner, while in some real-world applications, we cannot know the test data when training.
By contrast, our approach works in a more realistic inductive fashion that models cannot get information about test instances in the training phase.  
The only thing we can use is the unlabeled target-domain data, which can be different from the test data.

\section{Task Formulation}
In the few-shot classification, formally, we have two datasets: $\mathcal{D}_{meta-train}$ and $\mathcal{D}_{meta-test}$. 
These datasets contains a set of instances $(x, y)$ but $\mathcal{D}_{meta-train}$ and $\mathcal{D}_{meta-test}$ have their own label space that are disjoint with each other.
In the few-shot settings, $\mathcal{D}_{meta-test}$ is split into two parts: $\mathcal{D}_{support}$ and $\mathcal{D}_{query}$. 
If the support set contains $K$ instances for each of $N$ classes, this few-shot problem is called $N$-way-$K$-shot.
Usually, $K$ is really small, resulting in the poor performance when predicting $\mathcal{D}_{query}$.
Therefore, models should use $\mathcal{D}_{support}$ to predict $\mathcal{D}_{query}$ labels utilizing $\mathcal{D}_{meta-train}$.

For unsupervised domain adaptation in few-shot classification, $\mathcal{D}_{meta-train}$ and $\mathcal{D}_{meta-test}$ are sampled from different domains.
$\mathcal{D}_{meta-train}$ from the source domain and $\mathcal{D}_{meta-test}$ from the target domain.
We rename $\mathcal{D}_{meta-train}$ as $\mathcal{D}_{S}$.
To overcome domain discrepancy, an unlabeled target-domain dataset $\mathcal{D}_{UT}= \{ x_1, x_2, \ldots, x_{UT}\}$ is provided.
Our goal is to develop a model that acquires knowledge from the $\mathcal{D}_{S}$ and $\mathcal{D}_{UT}$, so that we can make predictions over $\mathcal{D}_{meta-test}$.

\section{Methodology}

\label{sec:methodology}

\begin{figure}[!tbp]
	\centering
	\includegraphics[width=1\linewidth]{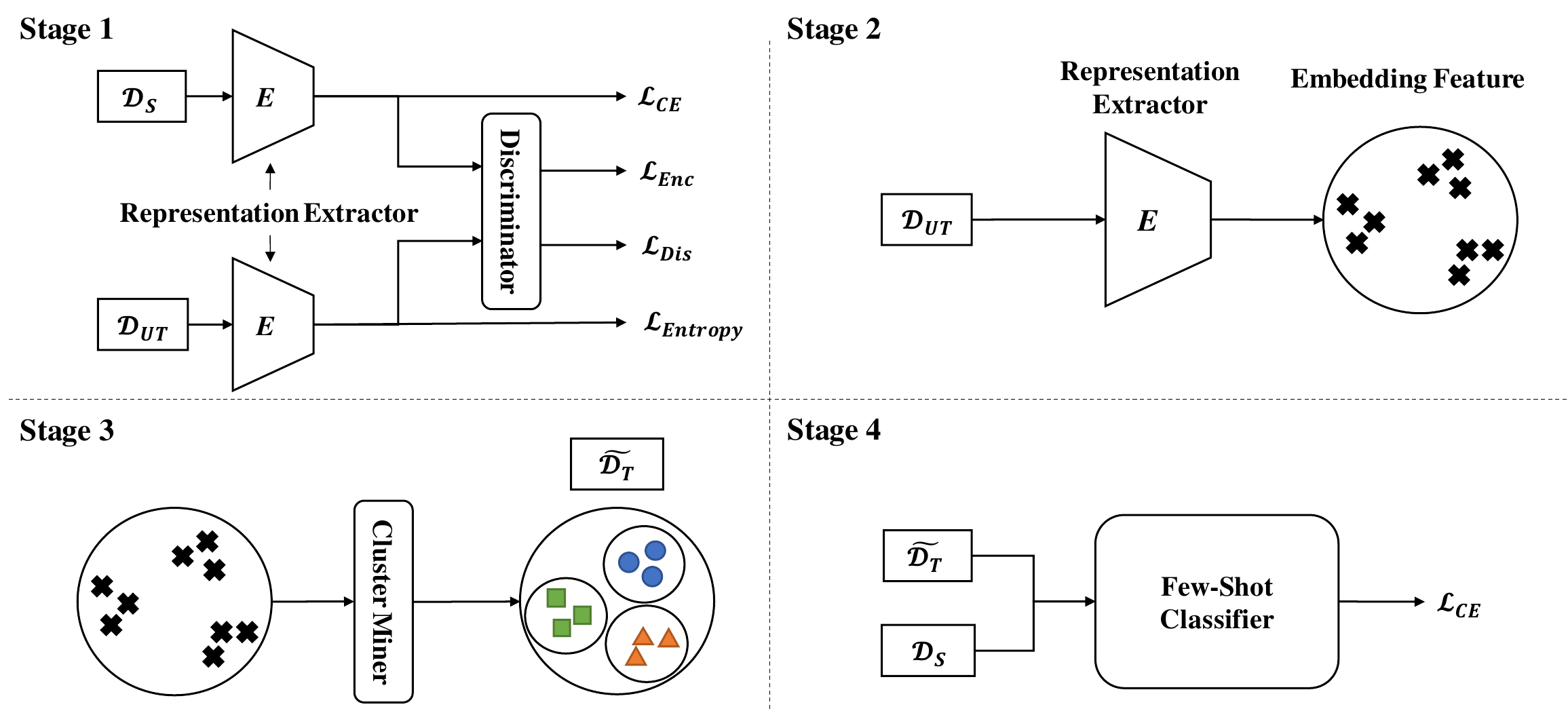}
	\caption{The overview of our DaFeC framework. In the first stage, we train a representation extractor based on our clustering promotion mechanism and then use it to extract features for unlabeled target-domain data. Next, all unlabeled target-domain instances are grouped with a cluster miner to generate pseudo-labels. In the last stage, the few-shot classifier is trained jointly with the target-domain pseudo-labeled data and the source-domain training data.}
	\label{img:framework}
\end{figure}

Figure \ref{img:framework} gives an over illustration of our framework, which operates in four stages as follows:
\begin{itemize}
	\item \textbf{Stage 1} Training the representation extractor with clustering promotion mechanism.
	\item \textbf{Stage 2} Extracting the features of unlabeled target-domain data.
	\item \textbf{Stage 3} Using the cluster miner to produce pseudo-labels for unlabeled target-domain data.
	\item \textbf{Stage 4} Training the few-shot classifier based on source-domain data and target-domain data with pseudo labels.
\end{itemize}

\subsection{DaFeC Framework}

\subsubsection{Representation Extractor}
The Representation Extractor $\bm{E}$ is used to extract features $\mathbf{x}$ for each input instance $x$.
Such operations is denoted as $\mathbf{x} = \bm{E}(x)$.
For the subsequent clustering, we use prototypical networks training method \cite{snell2017prototypical} and our proposed CPM (Details would be presented in Section \ref{sec:cpm}) to train our representation extractor.
Following~\cite{snell2017prototypical}, a Prototypical Vector representation for each class in $\mathcal{D}_{S}$ is generated, by averaging all the examples' representations of that label:
\begin{equation}
\mathbf{c}_i = \frac{1}{K} \sum^{K}_{j=1} \mathbf{x}^j_i, \quad i = 1, 2, \ldots, N, 
\end{equation}
where $\mathbf{c}_i$ refers to the prototype for class $y_i$ and $\mathbf{x}^j_i$ represents the embedding feature of the $j$-th instance of class $y_i$. Then the probability of each class for the query instance $x$ can be  computed as:
\begin{equation}
P(y=y_i | x) = \frac{\exp (-d(\bm{E}(x), \mathbf{c}_i))}{\sum^{N}_{j=1} \exp (-d(\bm{E}(x), \mathbf{c}_j))},
\end{equation}
where $d(\cdot, \cdot)$ is the Euclidean distance.
In the training phase, we expect to minimize the following objective function:
\begin{equation}
\min_{\theta} \mathcal{L}_{CE} = - \mathbb{E}_{x \in \mathcal{D}_{S}} [ \log P(y | x) ], 
\label{eqn:crossentropy}
\end{equation}
When updating the representation extractor with Equation \ref{eqn:crossentropy}, the Euclidean distance between each instance and the prototypical vector of its class could be reduced. As a result, instances with the same class get closer to their class centroid and away from other classes.
After training the representation extractor, we use it to embed all unlabeled target-domain data into embedding features for the next clustering stage.

\subsubsection{Cluster Miner}

Given the encoded features of all the target-domain instances in $\mathcal{D}_{UT}$ produced by the trained representation extractor, a cluster miner is deployed to group them into pre-defined $\tilde{N}$ distinct clusters.

Clustering has been widely studied and many approaches have been developed for a variety of circumstances. 
In our work, we focus on a standard clustering algorithm, $k$-means~\cite{macqueen1967some}. 
Same as the representation extractor, $k$-means get the centroid of each cluster by averaging all instances of that and use Euclidean distance to calculate the distance of every instance to their cluster centroid.
Therefore, the cluster results grouped by $k$-Means could reveal the cluster structure generated by the representation extractor better.
The subsequent cluster assignments are used as pseudo labels to guide the transformation of unlabeled $\mathcal{D}_{UT}$ to a pseudo-labeled dataset $\mathcal{\tilde{D}}_\mathcal{T}$, which is then merged with the source-domain training set $\mathcal{D}_{S}$ into a new training set $\mathcal{\tilde{D}}_{meta-train}=\{ \mathcal{D}_{S},\mathcal{\tilde{D}}_\mathcal{T} \}$.

\subsubsection{Few-shot Classifier}

The few-shot classifier is trained on $\mathcal{\tilde{D}}_{meta-train}$.
Because $\mathcal{\tilde{D}}_{meta-train}$ contains pseudo-labeled target-domain data, the performance of the classifier on the target domain could be improved.
Our proposed DaFeC is a generally applicable and model-agnostic framework, which means it is compatible with any existing few-shot classifier.
Following previous work~\cite{gao-etal-2019-fewrel}, we use Proto-CNN, Proto-BERT, BERT-PAIR for the classifier backbone to demonstrate the model-agnostic property of our framework.
All settings of these models are the same as the original paper.

\subsection{Clustering Promotion Mechanism}

\label{sec:cpm}

Generally, the few-shot classifier learns the information of the target domain by optimizing with pseudo labels created by the cluster miner.
While this seems reasonable, the inevitable label noise caused by the clustering procedure is ignored. 
Such noisy pseudo labels substantially hinder the model's capability to further improve the classification performance on the target domain. 
It is generally known that, as a typical machine learning algorithm, clustering depends heavily on the input representations, thus learning discriminative representations is fundamental to the high-confidence pseudo label generation.
In order to generate features with more discriminativeness, our framework further incorporates a novel \textbf{C}lustering \textbf{P}romotion \textbf{M}echanism (CPM) into the training process.
CPM is built on three components: Similarity Entropy Minimization, Adversarial Distribution Alignment, and Cosine Annealing Strategy.
We describe the details of all components below.

\subsubsection{Similarity Entropy Minimization}

Obviously, only if the features of similar instances are close together, the cluster miner can assign them the same pseudo label.
In order to promote this similarity without supervision, we introduce the Similarity Entropy Minimization (CPM-S) method.

We first compute the instance-wise distance vector $\mathbf{v} (\mathbf{x})$ for each target instance of $\mathcal{D}_{UT}$ as follows:
\begin{equation}
[\mathbf{v} (\mathbf{x}_i)]_j = \| \mathbf{x}_i - \mathbf{x}_j\|^2_2, \quad \mathbf{x}_i, \mathbf{x}_j \in \mathcal{D}_{UT}, \quad i \neq j,
\label{eqn:similarity}
\end{equation}
where $[\cdot]_j$ means the $j$-th element of a vector, $\| \cdot \|_2$ means the $l_2$ norm and $\mathbf{x}_i, \mathbf{x}_j$ are both from $\mathcal{D}_{UT}$.
To mine the latent cluster structure, we minimize the entropy of the normalized instance-wise distance vector $\mathbf{v} (\mathbf{x})$ for each target instance
\begin{equation}
\min_{\theta} \mathcal{L}_{Entropy} = \mathbb{E}_{x \sim \mathcal{D}_{UT}} [ H({\rm softmax}( \mathbf{v}(\mathbf{x}) / \tau)) ],
\label{eqn:entropy}
\end{equation}
where $H(\cdot)$ refers to the Shannon entropy over the softmax distribution, $\tau \in \mathbb{R}^{+}$ is a temperature scaling parameter of the softmax distribution to control the percentage of instances we expect the target data to be similar to. 
Too small $\tau$ sharpens the distribution as one-hot, resulting in several pair-wise clusters while setting too large $\tau$ can smooth the distribution to be uniform, making instances get close to dissimilar ones.

Different from the conventional class-level entropy minimization~\cite{DBLP:conf/nips/LongZ0J16} which calculate the entropy over the output logits of the classifier, our similarity entropy minimization over the instance-wise distance vector.
Through entropy minimization, the distribution of the instance-wise distance vector will be pushed away from the uniform distribution, which means that each instance is pushed to approach its similar ones and move away from other dissimilar samples. 
As a result, instances are encouraged to cluster together.

\subsubsection{Adversarial Distribution Alignment}

Naturally, training the representation extractor with Equation \ref{eqn:crossentropy}, the features of source-domain instances have been properly distributed into several distinct clusters.
Although in our few-shot scenario, the target domain does not share the same label space with the source domain, we still observe that they may have some similar classes that can be leveraged to promote target-domain instances to cluster together.
Recent efforts~\cite{Ganin2015DomainAdversarialTO,Tzeng2017AdversarialDD} have shown that adversarial training can align distributions of two domains, especially per-class distribution alignment.
Inspired by this, we introduce the Adversarial Distribution Alignment (CPM-A) method to promote the clustering of target-domain instances by mining similar classes across domains.
 
First, a domain discriminator $\bm{D}$ is built to accept features encoded by the representation extractor $\bm{E}$ and classify whether a data point is drawn from the source or the target domain. 
Thus, $\bm{D}$ is optimized according to a standard supervised loss where the labels indicate the origin domain, defined below:
\begin{equation}
\min_{\phi} \mathcal{L}_{Dis} = - \mathbb{E}_{x \sim \mathcal{D}_{S}} [\log \bm{D}_{\phi}(\bm{E}(x))] - \mathbb{E}_{x \sim \mathcal{D}_{UT}} [\log (1 - \bm{D}_{\phi}(\bm{E}(x)))]
\label{eqn:discriminator}
\end{equation}
Second, the representation extractor $\bm{E}$ playing as the generator is optimized to maximize $\mathcal{L}_{Dis}$, which updates $\bm{E}$ to generate features to confuse discriminator. This process can be reformulated as follow:
\begin{equation}
\min_{\theta} \mathcal{L}_{Enc} = - \mathbb{E}_{x \sim \mathcal{D}_{S}} [\log (1 - \bm{D}(\bm{E}_{\theta}(x)))] - \mathbb{E}_{x \sim \mathcal{D}_{UT}} [\log \bm{D}(\bm{E}_{\theta}(x))]
\label{eqn:encoder}
\end{equation}
Theoretically, by iterative optimization of Equation \ref{eqn:discriminator} and Equation \ref{eqn:encoder}, the domain discriminator $\bm{D}$ and the representation extractor $\bm{E}$ are alternated to reach the global optimality that $\bm{D}$ cannot distinguish between the features of source-domain and target-domain examples produced by $\bm{E}$.
Based on Equation~\ref{eqn:crossentropy}, the representation extractor is amended to encode instances of the source domain into cluster-distributed features. After the adversarial training phrase, the target-domain instances with similar ones in the source domain can be aligned with them as clusters.

\subsubsection{Cosine Annealing Strategy}

Unfortunately, CPM-S and CPM-A cannot work well by a simple multi-task learning strategy.
In the early training phase, the representation extractor has not learned well, so the produced features are crude and inaccurate. 
In this time, using CPM-S to promote clustering may make instances get close to dissimilar ones, resulting in undesirable clustering results.
With the training procedure going on, we can gradually increase the training weight of CPM-S to mine the latent cluster structure.
CPM-A helps to align distributions of source and target domain, but over-alignment could have a detrimental effect on target-domain class separation.
Since there still exist some target-domain classes completely different from classes in the source domain, the excessive alignment would mislead these classes to have inappropriate distributions for clustering, which hurts the quality of pseudo labels.
Previous work~\cite{han2018co} indicates that deep models would memorize easy instances first, and gradually adapt to hard instances as training epochs become large.
Thus in the training process, our representation extractor would first align distributions of similar classes between the two domains and then those of dissimilar classes.
Therefore, we can decrease the training weight of CPM-A gradually to allow model focus on the alignment of similar cross-domain classes and avoid unwanted over-alignment.

These observations motivate us to develop a Cosine Annealing Strategy (CPM-C) to adjust CPM-S and CPM-A weights in the training process.
Specifically, the overall loss function $\mathcal{L}$ is designed as the combination of  Equation~\ref{eqn:crossentropy}, Equation~\ref{eqn:encoder}, and Equation~\ref{eqn:similarity} as follows:
\begin{equation}
\min_{\theta} \mathcal{L} = \mathcal{L}_{CE} + (1-\lambda)\mathcal{L}_{Enc} + \lambda \mathcal{L}_{Entropy},
\label{eqn:loss}
\end{equation}
where $\lambda$ is the weighting parameter of $\mathcal{L}_{Entropy}$, which is designed to increase with the training epoch in the form of,
\begin{equation}
\lambda = \left\{ \begin{aligned}
&- \frac{\cos (\pi t / T) + 1}{2}, & t \le T\\
&1, & {\rm otherwise}
\end{aligned} \right. ,
\label{eqn:cosine}
\end{equation}
where $t$ is the current training epoch and $T$ denotes a pre-defined epoch annealing hyperparameter. 
In the early stage of the training procedure, $\mathcal{L}_{Enc}$ has a larger weight than $\mathcal{L}_{Entropy}$, which makes the representation extractor tend to learn transferable knowledge between domains to improve the ability of encoding target-domain instances. 
With the training procedure going on, the weight of $\mathcal{L}_{Entropy}$ will increase continually, so CPM-S can be encouraged to promote clustering for the target domain since the representation extractor has learned enough knowledge.
Compared to linear annealing, cosine function can pay more attention to CPM-A in the beginning and increase the weight of CPM-S more quickly.
We experimentally found that cosine annealing outperforms linear annealing with absolute gains (in classification accuracy) of 2\%-3\% under different settings.

\subsection{Overall Workflow}
In this section, we introduce the overall working procedure of our framework DaFeC. Algorithm \ref{alg:DaFeC} gives the scratch.

\begin{algorithm}[tb]
	\caption{DaFeC}
	\label{alg:DaFeC}
	\textbf{Input}: Labeled Source-domain Dataset $\mathcal{D}_{S}$ and Unlabeled Target-domain Dataset $\mathcal{D}_{UT}$\\
	\textbf{Output}: Few-shot Classifier $\mathbf{C}$
	\begin{algorithmic}[1] 
		\WHILE {not convergence}
		\STATE Sample $S$ and $Q$ from $\mathcal{D}_{S}$
		\STATE Sample $U$ from $\mathcal{D}_{UT}$
		\STATE Update representation extractor $\bm{E}$ with Equation \ref{eqn:crossentropy}
		\STATE Update discriminator $\bm{D}$ with Equation \ref{eqn:discriminator}
		\STATE Update representation extractor $\bm{E}$ with Equation \ref{eqn:encoder} and Equation \ref{eqn:entropy}
		\ENDWHILE
		\STATE Encode $\mathcal{D}_{UT}$ into feature representations $\{ \mathbf{x}_{{UT}} \}$ using $\bm{E}$.
		\STATE Run $k$-means on $\{ \mathbf{x}_{{UT}} \}$ to generate clusters $\mathcal{C}_{{UT}}$
		\STATE Assign each cluster in $\mathcal{C}_{{UT}}$ a pseudo label to construct pseudo-labeled dataset $\mathcal{\tilde{D}}_\mathcal{T}$
		\STATE Merge $\mathcal{\tilde{D}}_\mathcal{T}$ and $\mathcal{D}_{S}$ into $\mathcal{\tilde{D}}_{meta-train}$
		\STATE Train few-shot classifier $\mathbf{C}$ based on $\mathcal{\tilde{D}}_{meta-train}$
		\STATE \textbf{return} $\mathbf{C}$
	\end{algorithmic}
\end{algorithm}

Due to the size imbalance between $\mathcal{D}_S$ and $\mathcal{D}_{UT}$, we use the episodic paradigm proposed by \cite{vinyals2016matching} to train the representation extractor. 
In each iteration, $N$ classes are sampled from $\mathcal{D}_{S}$ randomly and each class will also randomly select $K$ instances as support instances. 
In this way, we can obtain the temporary support set $S$. 
And we choose other $M$ instances from the same $N$ classes to construct the temporary query set $Q$. 
Then the parameters of the representation extractor are optimized with Equation \ref{eqn:loss}. 
We use $S$ and $Q$ to calculate $\mathcal{L}_{CE}$. 
For the unlabeled target dataset $\mathcal{D}_{UT}$, we randomly sample $N \times K$ instances to construct the temporary unlabeled set $U$. 
$S$ and $U$ are encoded by the representation extractor into low-dimensional features. 
Then the discriminator takes these features as input and compute $\mathcal{L}_{Dis}$ and updates its parameters with Equation \ref{eqn:discriminator}. 
After that, we update the model weights of representation extractor following Equation \ref{eqn:loss}.
Once the representation extractor converges, we use it to encode all the instances in $\mathcal{D}_{UT}$ into embedding features $\{ \mathbf{x}_{{UT}} \}$. 
Next, we apply the $k$-means algorithm to mine latent cluster structure and assign them pseudo labels to construct pseudo-labeled target-domain dataset $\mathcal{\tilde{D}}_\mathcal{T}$, which is merged with the source-domain training set $\mathcal{D}_{S}$ into a new dataset $\mathcal{\tilde{D}}_{meta-train}$.
Finally, the few-shot classifier is trained based on $\mathcal{\tilde{D}}_{meta-train}$.

\section{Experiments}
\label{sec:experiments}

\subsection{Dataset and Metric}
We conduct experiments on the recently widely used benchmark FewRel 2.0 dataset introduced in \cite{gao-etal-2019-fewrel}, which is the currently largest unsupervised domain adaptation dataset for few-shot classification.
It consists of 44,800 labeled instances (64 classes and 700 instances per class) from Wikipedia (source domain) as the training set and 2500 labeled instances (25 classes and 100 instances per class) from Pubmed (target domain) as the test set. 
It also provides SemEval-2010 task 8 as the validation set (17 classes and 8,851 instances) and unlabeled PubMed data (2500 instances) for unsupervised domain adaptation.
This dataset focus on the relation classification task.
Each labeled example is a single sentence, annotated with a head entity, a tail entity, and their relation. 
The goal is to predict the correct relation between the head and tail. 

We investigate our experiments in four few-shot scenarios: 5-way-1-shot, 5-way-5-shot, 10-way-1-shot, 10-way-5-shot and report the mean and standard deviation of test accuracy according to the official evaluation scripts \footnote{https://thunlp.github.io/2/fewrel2\_da.html}.

\subsection{Implementation Details}

Following \cite{zeng2014relation}, we implement the representation extractor $\bm{E}$ based on a convolutional neural network to encode sentences for relation classification.
The window size of CNN is set to 3, and the number of filters is 230.
The discriminator $\bm{D}$ is implemented as a two-layer feed forward neural network.
We use the 50 dimension Glove embeddings~\cite{pennington2014glove} to initialize word embeddings. 
Following~\cite{zeng2014relation}, we also concatenated the input word embeddings with 5-dimensional position embeddings.
The model is trained using stochastic gradient descent with the learning rate of 0.1.
$\tilde{N}$ (the number of clusters), $\tau$ (the temperature scaling parameter) and $T$ (the epoch annealing parameter) are set as 10, 2 and 6000, respectively.
All the hyper-parameters are tuned on the validation set.
We run all experiments using PyTorch 1.1.0 on the Nvidia Tesla V100 GPU.

\subsection{Baselines}

We compare our model against the following 5 models proposed in \cite{gao-etal-2019-fewrel}:

\begin{itemize}
	\item \textbf{Proto-CNN} is a prototypical network using CNN\cite{zeng2014relation} as the encoder. 
	\item \textbf{Proto-BERT} is also a prototypical network but it uses BERT~\cite{DBLP:conf/naacl/DevlinCLT19} as its encoder. 
	\item \textbf{Proto-CNN-ADV} is straightforward to combine traditional domain adaptation technique, adversarial training, with few-shot model, Proto-CNN.
	\item \textbf{Proto-BERT-ADV} like Proto-CNN-ADV, simply utilizes adversarial training technique to augment Proto-BERT 
	\item \textbf{BERT-PAIR} pairs each query instance with all the supporting instances, and send the paired sequence to the BERT sequence classification model, which is the state-of-the-art on the FewRel 2.0 dataset.
\end{itemize}

\subsection{Results}

\begin{table}[!tb]
	\centering
	\begin{tabular}{lcccc}  
		\toprule
		Model  & 5-Way-1-Shot & 5-Way-5-Shot & 10-Way-1-Shot & 10-Way-5-Shot\\
		\midrule
		Proto-CNN 			& 35.09 $\pm$ 0.10 & 49.37 $\pm$ 0.10 & 22.98 $\pm$ 0.05 & 35.22 $\pm$ 0.06 \\
		Proto-BERT 			& 40.12 $\pm$ 0.19 & 51.50 $\pm$ 0.29 & 26.45 $\pm$ 0.10 & 36.93 $\pm$ 0.01 \\
		BERT-PAIR 			& 56.25 $\pm$ 0.40 & 67.44 $\pm$ 0.54 & 43.64 $\pm$ 0.46 & 53.17 $\pm$ 0.09 \\
		\midrule
		Proto-CNN-ADV 		& 42.21 $\pm$ 0.09 & 58.71 $\pm$ 0.06 & 28.91 $\pm$ 0.10 & 44.35 $\pm$ 0.09 \\
		Proto-BERT-ADV 		& 41.90 $\pm$ 0.44 & 54.74 $\pm$ 0.22 & 27.36 $\pm$ 0.50 & 37.40 $\pm$ 0.36 \\
		\midrule
		DaFeC+Proto-CNN 	& 48.58 $\pm$ 0.65 & 65.80 $\pm$ 0.44 & 35.53 $\pm$ 0.67 & 52.71 $\pm$ 0.54\\
		DaFeC+Proto-BERT 	& 46.39 $\pm$ 0.68 & 56.32 $\pm$ 0.84 & 32.09 $\pm$ 0.98 & 40.53 $\pm$ 0.75\\
		DaFeC+BERT-PAIR 	& \textbf{61.20 $\pm$ 0.91} & \textbf{76.99 $\pm$ 0.82} & \textbf{47.63 $\pm$ 1.01} & \textbf{64.79 $\pm$ 0.77}\\
		\bottomrule
	\end{tabular}
	\vspace{1mm}
	\caption{Accuracies (\%) of different models on the FewRel 2.0 test set with domain adaptation. Bold marks the highest number among all models. All the results of baseline models are quoted directly from~\cite{gao-etal-2019-fewrel}. ``DaFeC+'' denotes our proposed method.}
	\label{tab:accuracies}
\end{table}

Table \ref{tab:accuracies} reports the results of our methods (DaFeC+Proto-CNN, DaFeC+Proto-BERT and DaFeC+BERT-PAIR) against other baseline methods. 
From the results, we can observe that: 
(1) Over the previous state-of-the-art method BERT-PAIR, DaFeC+BERT-PAIR achieves substantial improvements of 4.95\%, 9.55\%, 3.99\% and 11.62\% on four few-shot settings respectively, which confirms the effectiveness and rationality of our proposed training framework.
(2) Besides DaFeC+BERT-PAIR, both DaFeC+Proto-CNN and DaFeC+Proto-BERT also exceed Proto-CNN and Proto-BERT significantly. 
The accuracy of DaFeC+Proto-CNN and DaFeC+Proto-BERT increase 7.11\% and 3.48\% on average compared to Proto-CNN and Proto-BERT. 
This demonstrates the model-agnostic property of our framework.
(3) Our models with DaFeC clearly perform better than the Proto-CNN-ADV and Proto-BERT-ADV, showing that naive applying UDA to few-shot classification is not as effective as our specifically designed framework. 
(4) The standard deviations of DaFeC+Proto-CNN, DaFeC+Proto-BERT and DaFeC+BERT-PAIR are slightly larger than the original models because of the pseudo-labeled noise. 
However, these models still outperform the original ones even considering the worst performance. 

\section{Analyses}

\subsection{Ablation Study}

To study the contribution of each component in our framework, we run an ablation study (see also Table \ref{tab:ablation}). From these ablations, we find that:
 (1) Removing the pseudo-labeled target-domain data hurts the result by 4.73\%, 8.36\%, 4.80\% and 9.97\% in four scenarios, respectively, which indicates that training the network with clustering-generated pseudo labels is vital for domain adaptation.  
 (2) By introducing the similarity entropy minimization method (CPM-S), we can mine the latent cluster structure of unlabeled target-domain instances, which is beneficial to high-quality pseudo label generation.
 (3) When we remove CPM-A, the score drops by 3.81\% on average, which demonstrates the effectiveness of adversarial distribution alignment over similar classes across domains.
 (4) When we fix the coefficients in Equation~\ref{eqn:loss} rather than use the cosine annealing strategy to adjust them, the performance declines extremely and is not on par even with only using CPM-S and CPM-A, which powerfully proves that CPM-S and CPM-A cannot work properly without CPM-C.

\begin{table*}[!tb]
	\centering
	\begin{tabular}{lcccc}  
		\toprule
		Model  & 5-Way-1-Shot & 5-Way-5-Shot & 10-Way-1-Shot & 10-Way-5-Shot\\
		\midrule
		DaFeC+BERT-PAIR				& 61.00 & 76.83 & 46.00 & 65.27\\
		\quad - Pseudo-labeled Data	& 56.27 & 68.47 & 41.20 & 55.30\\
		\quad - CPM-S 				& 59.07 & 72.03 & 43.13 & 59.43\\
		\quad - CPM-A 				& 58.13 & 71.43 & 43.10 & 60.56\\
		\quad - CPM-C 				& 57.50 & 69.40 & 42.63 & 57.43\\
		\bottomrule
	\end{tabular}
	\vspace{1mm}
	\caption{An ablation study of our proposed framework on the FewRel 2.0 dataset.}
	\label{tab:ablation}
\end{table*}

\subsection{Effectiveness of CPM}

\begin{figure}[tb]
	\centering
	\subfigure[Davies-Boulbin Index]{
		\begin{minipage}[h]{0.5\linewidth}
			\centering
			\includegraphics[scale=0.4]{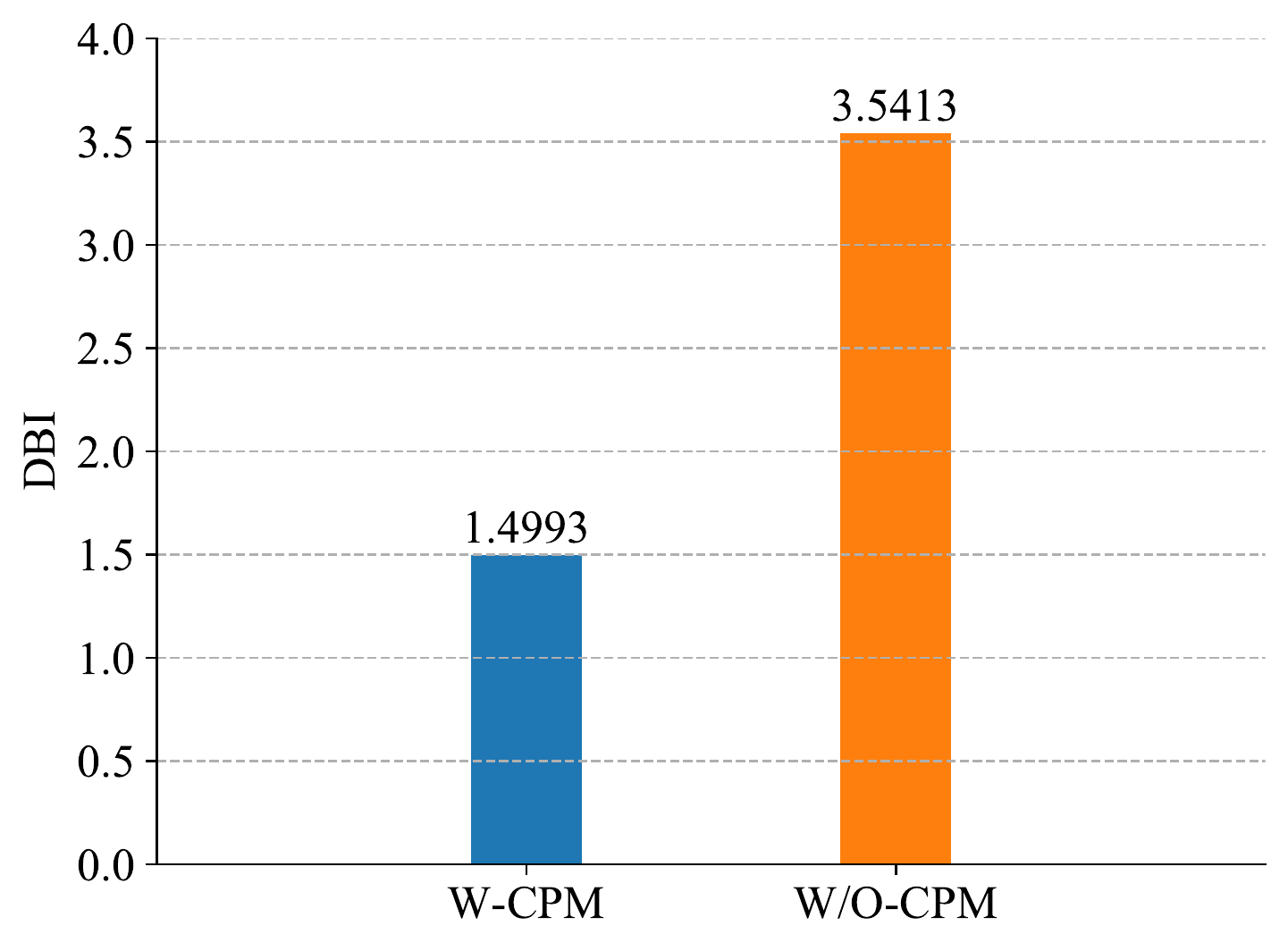}
			\label{fig:dbi}
		\end{minipage}%
	}%
	\subfigure[Fowlkes and Mallows Index]{
		\begin{minipage}[h]{0.5\linewidth}
			\centering
			\includegraphics[scale=0.4]{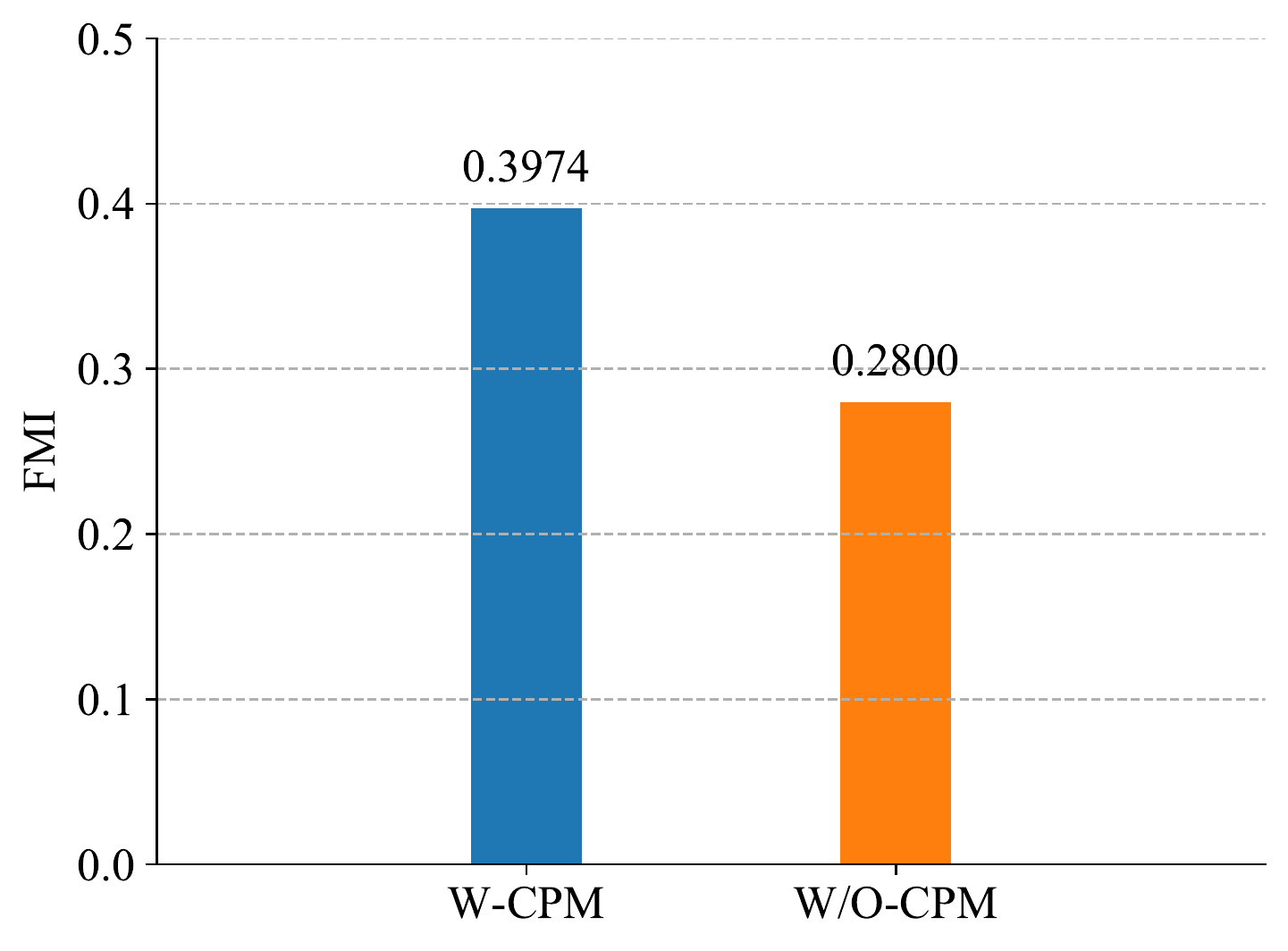}
			\label{fig:fmi}
		\end{minipage}%
	}%
	\caption{Validation analysis of CPM. The Davies-Boulbin Index (DBI) is used to represent the tightness in a cluster (lower is better) and the Fowlkes and Mallows Index (FMI) is employed to indicate the clustering accuracy (higher is better).}
	\label{fig:validity}
\end{figure}

The effectiveness of the learned discriminative feature representations through CPM can be investigated quantitatively and qualitatively.

%
%

%
In order to measure the cohesiveness of intra-clusters and the separation of inter-clusters, we calculate Davies-Boulbin Index (DBI)~\cite{davies1979cluster} on the unlabeled target-domain data.
We first train two representation extractor: (1) one trained with CPM, named \textit{W-CPM}, (2) the other trained without CPM, named \textit{W/O-CPM}, and use them to encode the unlabeled target-domain instances from the training set respectively.
Then the DBI value is calculated based on the cluster results of these features, and the lower, the better.
From Figure \ref{fig:dbi}, we observe that W-CPM yields a considerably lower DBI value (1.4993) compared with \textit{W/O-CPM} (3.5413).

To examine the accuracy of clustering, we first generate pseudo labels for unlabeled target-domain training data using \textit{W-CPM} and \textit{W/O-CPM}, respectively.
Then the Fowlkes and Mallows Index (FMI)~\cite{fowlkes1983method} score can be obtained by comparing the pseudo labels with the ground truth labels~\footnote{Note that FewRel 2.0 provides ground-truth labels for partial target-domain training data, but researchers are forbidden to use these labels in the training process for unsupervised domain adaption. And we only use them for experimental analysis. }, and the higher, the better.  
The results shown in Figure \ref{fig:fmi} suggest that CPM has indeed improved the clustering effect by increasing the FMI score from 0.28 to 0.3974.

%
%

\begin{figure}[tb]
	\centering
	\subfigure[Representations of W/O-CPM]{
		\begin{minipage}[h]{0.5\linewidth}
			\centering
			\includegraphics[scale=0.45]{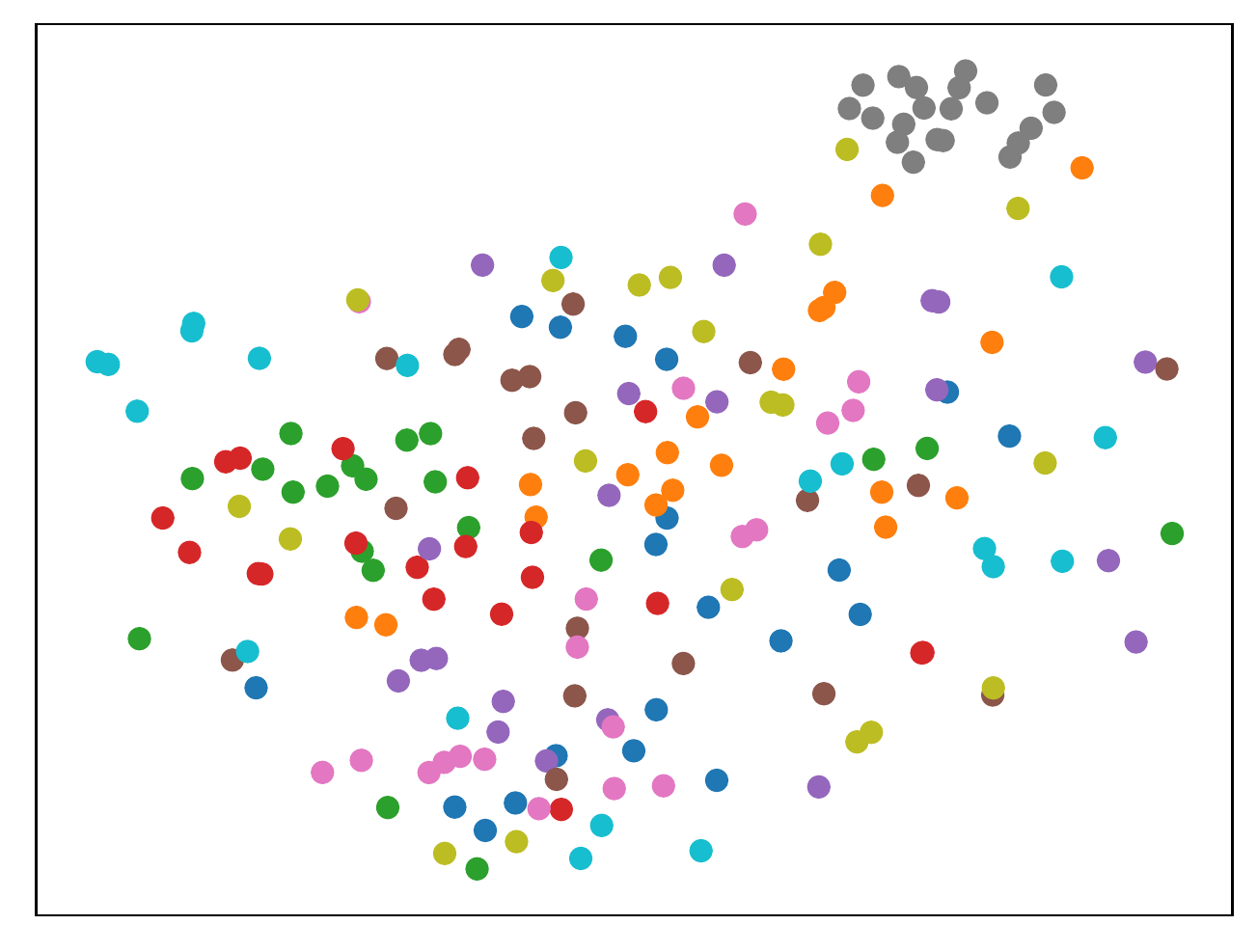}
			\label{fig:non-fea}
		\end{minipage}%
	}%
	\subfigure[Representations of W-CPM]{
		\begin{minipage}[h]{0.5\linewidth}
			\centering
			\includegraphics[scale=0.45]{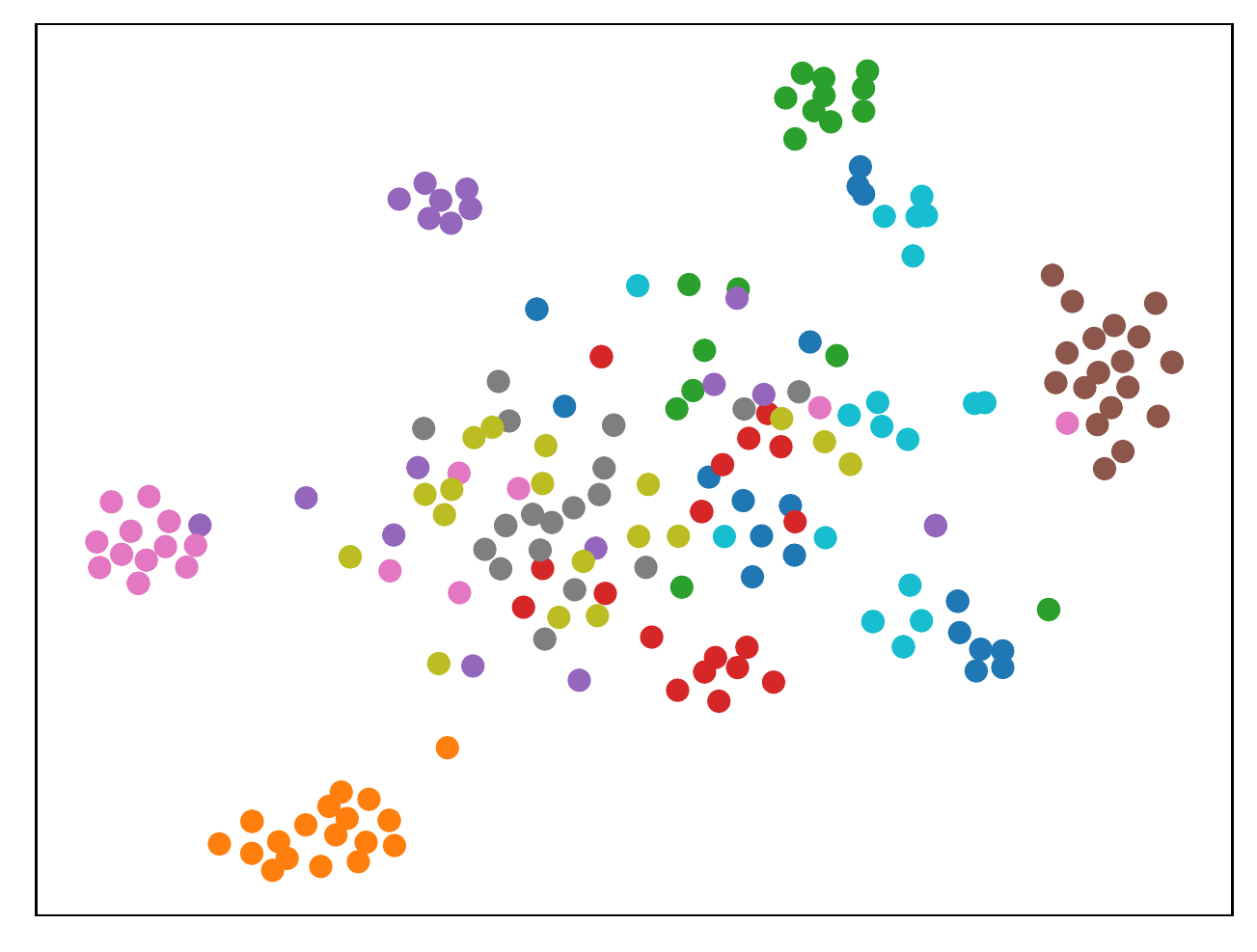}
			\label{fig:all-fea}
		\end{minipage}%
	}%
	\caption{A t-SNE plot of the computed feature representations of target-domain instances in the FewRel 2.0 training set. Node colors denote classes.The difference between Figure \ref{fig:all-fea} and Figure \ref{fig:non-fea} is whether the representation extractor is trained with CPM or not.}
	\label{fig:tsne}
\end{figure}

In addition, we visualize the features with t-SNE projected onto 2D embedding space.
Specifically, we sample 20 instances for each of 10 classes from the target-domain training data with ground truth labels and extract their features using two representation extractors, \textit{W-CPM} and \textit{W/O-CPM}.
Figure \ref{fig:tsne} provides the visualization of the t-SNE-transformed feature representations.
We can observe that for the model without trained with CPM, the features actually are mixed and the points with the same classes are distributed in different places. Thus, the pseudo labels generated by the cluster miner may have mush noise. 
While for the model trained with CPM, the representation exhibits discernible clustering in the projected 2D space. Therefore, the cluster results of the cluster miner could have higher quality.

We draw the conclusion that CPM can enhance the discriminativeness of target-domain feature representations and make those instances distribute as clusters. 
Therefore, when the cluster miner generates pseudo labels for unlabeled target-domain data according to the features encoded by the model trained with CPM, the pseudo labels could have higher confidence, which may provide more useful target-domain information to improve the domain adaption ability of the few-shot classifier.

\subsection{Error Analysis}
Although our method achieves state-of-the-art results, we still observe some phenomena which could cause failures.
From Figure \ref{fig:all-fea}, we find that five kinds of colored instances, red (\textit{ingredient\-\_of}), grey (\textit{causative\-\_agent\-\_of}), yellow (\textit{classified\_as}), cyan (\textit{gene\_plays\-\_role\-\_in\_process}) and blue (\textit{biological\_process\-\_involves\_gene\_product}), fail to show the cluster structure.
It reveals that even trained with CPM, the representation extractor still cannot produce discriminative features for all the target-domain data.
In other words, some instances belonging to different classes may have similar representations. 
As a result, the pseudo labels produced by the cluster miner inevitably contain noise that some instances from different classes are assigned the same pseudo label, while other instances from the same class could have distinct labels, which limits the few-shot classifier as its performance on the target domain is determined by the quality of the pseudo labels.

\section{Conclusion}
In this paper, we study the problem of inductive unsupervised domain adaption in the few-shot classification.
We first train a representation extractor with the Clustering Promotion Mechanism. 
It uses Similarity Entropy Minimization to promote clustering and Adversarial Distribution Alignment to align similar class distribution across domains.
Two methods are combined by the proposed Cosine Annealing Strategy. 
The representation extractor is used to encode unlabeled target-domain data into features, which are passed to a $k$-means cluster miner to generate pseudo labels.
Finally, we utilize pseudo-labeled target-domain data and labeled source-domain data to train the few-shot classifier. 
Experimental results demonstrate that our approach achieves new state-of-the-art on FewRel 2.0 dataset.
In the future, we will work on how to reduce the noise of pseudo labels to improve the domain adaption performance.

\section*{Acknowledgements}

We would like to thank reviewers for their insightful comments.
This work is supported in part by the National Key Research and Development Program of China (grant No. 2016YFB0801003), the Strategic Priority Research Program of Chinese Academy of Sciences (grant No. XDC02040400) and grant No. BMKY2019B04-1.

%
%

%
\end{document}